\documentclass{Interspeech}

\interspeechcameraready

\usepackage{amssymb}
\usepackage{pifont}
\usepackage{eso-pic}

\newcommand{\interspeechLeftHeader}{
  \AddToShipoutPictureFG*{
    \AtTextUpperLeft{
      \hspace*{-2mm}
      \raisebox{1.0cm}[0pt][0pt]{
        \parbox{\textwidth}{
          \raggedright\small
          Published in \textit{Proceedings of Interspeech 2025}, pp. 276--280.\\
          DOI: \url{https://doi.org/10.21437/Interspeech.2025-2792}\\
          Proceedings: \url{https://www.isca-archive.org/interspeech_2025/}
        }
      }
    }
  }
}

\newcommand{\cmark}{\ding{51}}
\newcommand{\xmark}{\ding{55}}

\title{Towards Multi-Level Transcript Segmentation: \\LoRA Fine-Tuning for Table-of-Contents Generation}

\author{Steffen}{Freisinger}
\author{Philipp}{Seeberger}
\author{Thomas}{Ranzenberger}
\author{Tobias}{Bocklet}
\author{Korbinian}{Riedhammer}

\affiliation[nocounter]{}{Technische Hochschule Nürnberg}{Germany}
\email{firstname.lastname@th-nuernberg.de}

\usepackage{comment}

\begin{document}

\interspeechLeftHeader

\maketitle

\begin{abstract}
Segmenting speech transcripts into thematic sections benefits both downstream processing and users who depend on written text for accessibility. We introduce a novel approach to hierarchical topic segmentation in transcripts, generating multi-level tables of contents that capture both topic and subtopic boundaries. We compare zero-shot prompting and LoRA fine-tuning on large language models, while also exploring the integration of high-level speech pause features. Evaluations on English meeting recordings and multilingual lecture transcripts (Portuguese, German) show significant improvements over established topic segmentation baselines. Additionally, we adapt a common evaluation measure for multi-level segmentation, taking into account all hierarchical levels within one metric.
\end{abstract}

\section{Introduction}

Topic segmentation of spoken content is essential for many speech technology applications, especially when dealing with lengthy or multi-speaker recordings such as lectures, meetings, and seminars \cite{Park-2023, ghinassi-2023, Soares-2018}.
Unlike written texts, which are usually structured into chapters or sections (e.g., books, papers, etc.), automatically generated transcripts of spoken language lack this semi-structural information.
Despite advances in automatic speech recognition (ASR), the absence of structure makes it difficult for users to find specific information, and downstream tasks such as information retrieval (IR) and retrieval augmented generation (RAG) rely on meaningful segmentation.

Most previous work in spoken document segmentation has explored traditional linear segmentation methods, where the speech transcripts are divided into segments based on the temporal dimension.
However, topics often exist on several levels of granularity, and the exclusively temporal view completely ignores these compositional structures.
For example, a broad topic like "Project Requirements" may encompass more specific subtopics such as "User Interface" and "Testing Strategies". 
Treating these subtopics as equivalent to broader themes disrupts the inherent hierarchy, resulting in a loss of structural clarity.
A hierarchical approach not only improves navigation for all users, but also benefits those with hearing impairments by providing a clear, multi-level structure in written form.
This fosters inclusive access to spoken content, where transcripts serve as a primary means of information intake.

Large Language Models (LLMs) such as LLaMA\footnote{https://arxiv.org/abs/2307.09288}, Qwen\footnote{https://arxiv.org/abs/2412.15115}, and GPT-4\footnote{https://arxiv.org/abs/2303.08774} demonstrate strong language understanding and have led to significant advances in tasks such as summarization \cite{Shang-2024,Kang-2024} and report generation \cite{reddy2024smartbook}, which require an understanding of coherent content.
Furthermore, their ability to perform zero-shot and in-context learning has unlocked new capabilities by leveraging task descriptions alongside large-scale pre-training and instruction-following data.
However, applying LLMs to generate table of contents (ToC) in spoken content remains underexplored.

This work is inspired by recent successes in using LLMs for related tasks \cite{Shang-2024,Kang-2024,reddy2024smartbook}.
We propose an approach to bridge the gap between the potential of LLMs for ToC generation and hierarchical topic segmentation.
Our goal is to automatically generate a hierarchical topical structure for spoken language content\footnote{Code and prompts: https://github.com/steffrs/toc-llm}.

Our main contributions are summarized as follows:
\begin{itemize}
\item We present a novel approach to the task of hierarchical topic segmentation of transcripts based on advances in summarization using LLMs.
\item We demonstrate that our method outperforms existing segmentation baselines and evaluate the linear segmentation task on three datasets.
\item In terms of multi-level segmentation, we adapt a common linear segmentation metric, so that it considers all hierarchical levels.
\end{itemize}

\section{Related Work}
\label{sec:related-work}

\subsection{Spoken Document Segmentation}
Spoken document segmentation applies to various domains, including lecture videos~\cite{Soares-2018, Chand-2021}, meetings~\cite{ICSI-2003, AMI-2005}, radio programs~\cite{ghinassi-2023, Berlage-2020}, dialogues~\cite{Zhang-2019}, podcasts~\cite{Clifton-2020, Garmash-2023}, and YouTube videos~\cite{Retkowski-2024}.
Most work focuses on segmenting textual transcripts, with a line of research incorporating acoustic features.
TextTiling~\cite{Hearst-1997} determines topic shifts via lexical deviations across sentences and can be extended with pre-trained sentence embeddings~\cite{Solbiati-2021} or randomly drawn hyperdimensional vectors~\cite{Park-2023}. 
Typical supervised methods treat segmentation as a sequence tagging task, labeling sentences as \textit{topic change} or \textit{no change}, using LSTM- and Transformer-based models \cite{Zhang-2019, Koshorek-2018}.
Other work frame it as a classification problem and determine whether a given sentence boundary presents a topic change, given its left and right context~\cite{Lukasik-2020, Ghosh-2022}.
Beyond text, other approaches incorporate acoustic cues such as pauses, pitch, volume, and speaker changes~\cite{Chand-2021, Soares-2019} as well as embeddings from pre-trained audio models~\cite{ghinassi-2023, Berlage-2020}.
Despite the progress made in segmenting spoken documents, the focus is primarily on the flat level and ignores hierarchical structures.

\subsection{Hierarchical Topic Segmentation}
Unlike traditional topic segmentation, hierarchical topic segmentation identifies topic shifts at multiple levels.
Early work in this area combined lexical cohesion measures with conversational features to detect both top-level and subtopic boundaries in multiparty dialogues \cite{Hsueh-2006}.
Later approaches leveraged Bayesian multi-scale models and affinity propagation to construct nested topic structures in text~\cite{Eisenstein-2009, Kazantseva-2014}.
Hierarchical methods have also been explored in spoken dialogue summarization. In \cite{Li-2021}, the authors proposed a two-stage system integrating ASR with text-based summarization, employing specialized segmentation and merging algorithms to process long conversational data. 
Although not exclusively a segmentation framework, their approach highlights the value of multi-level organization for enhanced readability.

\subsection{Table-of-Contents Generation}

A specialized form of topic segmentation is ToC generation, which integrates segmentation with title generation.
\cite{Branavan-2007} use a hierarchical discriminative model to generate ToCs for textbooks.
Their method combines local and global models for candidate title generation and candidate title selection, respectively.
Here, the global model aims to enforce the coherence of the ToC structure.
Similarly, \cite{Erbs-2013} investigate automatic hierarchy identification, leveraging linguistic features and supervised classification on diverse corpora such as Wikipedia and literary works.
To address data sparsity, \cite{Nguyen-2009} employ semi-supervised learning to integrate word clusters into structured models in order to enhance segmentation for long texts.
Considering speech transcripts, \cite{Ohtsuki-2004} apply hierarchical clustering on conceptual word vectors to extract a layered topic structure in meeting transcripts.
In contrast, PODTILE \cite{Azin-2024}  introduces a fine-tuned transformer for generating podcast chapters with titles but does not incorporate hierarchical structures.
While early research explored this direction, end-to-end generation of ToC using LLMs remains relatively underexplored in spoken document segmentation.

\section{Method}

\subsection{Problem Definition}
\label{sec:problem_definition}
Let \(D\) be a document of \(n\) sentences \(\{s_1, \ldots, s_n\}\). A \textit{linear segmentation} partitions \(D\) into a single set of contiguous segments \(\{S_1, \ldots, S_m\}\), such that \(\bigcup_{i=1}^{m} S_i = D\) and no two segments overlap. We extend this to a \textit{hierarchical segmentation} across \(L\) levels, where each level \(l\) also defines a full partition of \(D\). Every boundary at level \(l\) persists at level \(l+1\). Formally, if \(S_i^l\) is a segment at level \(l\), it may stay intact at level \(l+1\), or split into smaller contiguous sub-segments \(\{S_{i}^l\}\to\{S_{i,1}^{l+1}, S_{i,2}^{l+1}, \dots\}\). This produces nested structures, from coarse top-level topics at \(l=1\) down to finer subtopics at \(l=L\). For our linear segmentation experiments, we consider only the lowest level \(L\) in isolation.

\subsection{ToC-LLM: Table-of-Contents Generation}
\begin{figure}[t]
  \centering
  \includegraphics[width=\linewidth]{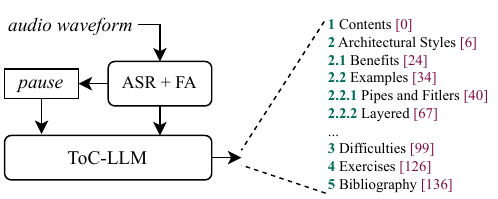}
  \caption{Overview of our approach with an example from \textsc{Videoaula} (translated): First, we generate transcripts using an ASR system and apply forced alignment (FA) to synchronize the transcripts with the audio. Then, we extract inter-sentence pause durations and prompt the LLM along with the text. The output is a table of contents that shows the topic boundaries.}
  \label{fig:framework}
\end{figure}

We present a method that applies LLMs to produce a multi-level table of contents directly from a speech transcript (cf. Figure~\ref{fig:framework}). Rather than predicting only boundary positions, our approach generates a structured outline, allowing each segment to be labeled with its hierarchical level. This is achieved through a system prompt instructing the model to output a table of contents for the entire transcript. Each line of the input contains exactly one sentence, labeled with its corresponding index \cite{Fan-2024, Yu-2023}. The final output uses dotted numbering (e.g., \texttt{2.2.1}) to encode the segment’s level, along with a topic name and the associated sentence index. Figure~\ref{fig:framework} illustrates an example of the resulting structure.

\subsection{Infusing Acoustic Cues}
To explore the impact of acoustic cues on topic segmentation, we optionally incorporate inter-sentence pause durations into the transcript text. Past research indicates that pauses serve as a valuable indicator of topic boundaries \cite{ghinassi-2023, Soares-2018, freisinger2023_slate}. In our setup, each sentence line can include a pause annotation (e.g., \texttt{42 (pause=0.62s): Now let's continue with...}), allowing the model to see these timing gaps. Because of the precise alignment performed during preprocessing (cf. Section~\ref{sec:datasets}), we obtain pause intervals directly from the sentence boundaries without additional audio processing. This optional feature aims to leverage high-level acoustic information to refine boundary detection.

\subsection{Modeling and Implementation}
Our primary model is \textit{Mistral Nemo}\footnote{https://mistral.ai/news/mistral-nemo}, a 12-billion-parameter LLM trained on multilingual data. It supports a 128k-token context window, making it suitable for long-form transcripts in different languages. For comparison, we also employ \textit{Qwen2.5-7B}\footnote{https://arxiv.org/abs/2412.15115}, a 7.61-billion-parameter model with the same context length and multilingual coverage. Both models have been instruction-tuned to better handle natural language queries. We refer to these models as \textsc{ToC-Nemo} and \textsc{ToC-Qwen}.

We use 4-bit quantized checkpoints\footnote{\textit{unsloth/Qwen2.5-7B-Instruct-unsloth-bnb-4bit}}\footnote{\textit{unsloth/Mistral-Nemo-Instruct-2407-bnb-4bit}} 
from the \textit{unsloth}\footnote{https://github.com/unslothai/unsloth} library to reduce memory usage and accelerate inference. To explore domain adaptation, we compare a zero-shot scenario (no fine-tuning) with LoRA-based fine-tuning \cite{Hu-2022}. LoRA freezes the main model weights and updates a set of low-rank adapters.
In combination with 4-bit quantization, this has shown to significantly reduce the memory footprint and training time, while preserving performance comparable to full-precision fine-tuning ~\cite{Dettmers-2023}.
In both zero-shot and fine-tuned configurations, we maintain the same prompt format to ensure a fair comparison.

\section{Experiments}

\renewcommand{\arraystretch}{1.15}
\begin{table*}[t!]
  \caption{
  Comparison of methods for different datasets.
  We report mean ± stddev across 100 bootstrapped test sets (\textsc{AMI}) and mean ± stddev on leave-one-speaker-out cross-validation sets (\textsc{Videoaula}, \textsc{LectureDE}). \cmark~indicates fine-tuning on respective target train set and \xmark~refers to the zero-shot setting. The best scores in each group are highlighted in bold.
  }
  \label{tab:results}
  \centering
  \resizebox{\linewidth}{!}{
  \begin{tabular}{l c c c | c c | c c}
    \toprule
    & 
    \multicolumn{1}{c}{} & \multicolumn{2}{c}{\textbf{\textsc{AMI}}} 
    & \multicolumn{2}{c}{\textbf{\textsc{Videoaula}}} 
    & \multicolumn{2}{c}{\textbf{\textsc{LectureDE}}} \\
    \cmidrule(lr){3-4}
    \cmidrule(lr){5-6}
    \cmidrule(lr){7-8}
    \textbf{Methods} 
    & \multicolumn{1}{c}{\textbf{Tuning}}
    & \multicolumn{1}{c}{\(F_1\uparrow\)}
    &  \multicolumn{1}{c}{\(B\uparrow\)}
    & \multicolumn{1}{c}{\(F_1\uparrow\)}
    &  \multicolumn{1}{c}{\(B\uparrow\)}
    & \multicolumn{1}{c}{\(F_1\uparrow\)}
    & \multicolumn{1}{c}{\(B\uparrow\)}
    \\
    \midrule
    \textsc{TT-BERT} \cite{Solbiati-2021}  &  \xmark   & 7.25 ± 1.47 & 6.71 ± 0.97 & 9.42 ± 2.45 & 8.73 ± 2.87 & 6.94 ± 4.43 & 8.31 ± 2.55 \\
    \textsc{SegmentLLM ChatGPT} \cite{Fan-2024} &   \xmark  &  \textbf{10.35} ± 2.14    &   \textbf{7.97} ± 1.31   &  2\textbf{7.57} ± 9.27   &  \textbf{22.29} ± 8.32   &   -  &  - \\
    \textsc{SegmentLLM Nemo} cf. \cite{Fan-2024}  &  \xmark   &   5.42 ± 1.44   &  5.79 ± 1.02   &  4.92 ± 2.64   &   4.68 ± 2.18   &  9.87 ± 9.17   & 11.37 ± 10.22  \\
    \textsc{ToC-Nemo}  & \xmark  &   \underline{7.36} ± 1.51 & \underline{7.68} ± 1.07  &   \underline{12.27} ± 6.45 &  \underline{14.06} ± 5.80 & 20.55 ± 5.97  & 17.04 ± 4.98     \\
    \textsc{ToC-Nemo + Pause}   & \xmark & 6.50 ± 2.23 & 7.05 ± 1.72  &  9.90 ± 3.14 & 12.4 ± 3.79 & 17.5 ± 7.64  &  15.79 ± 6.26   \\
    \midrule
    \textsc{MiniSeg} \cite{Retkowski-2024} &  \cmark   & 18.61 ± 2.54 & 14.90 ± 1.68 & 38.72 ± 7.80 & 28.28 ± 4.42 & 23.86 ± 11.18 & 19.21 ± 8.11 \\
    \textsc{CS-BERT} \cite{Lukasik-2020}  &  \cmark   & 20.50 ± 2.71 & 15.77 ± 1.78 & 47.77 ± 8.47 & 34.22 ± 6.09 & 31.08 ± 5.99 & 22.88 ± 6.37 \\
    \textsc{ToC-Nemo}  &  \cmark  & \underline{26.55} ± 2.92 & \underline{22.79} ± 2.40 & \underline{64.26} ± 7.86 & \underline{51.94} ± 8.10 & \underline{58.47} ± 15.86 & \underline{49.50} ± 17.04 \\
    \textsc{ToC-Nemo + Pause} &  \cmark  &  \textbf{30.34} ± 3.28 &  \textbf{24.81} ± 2.30 & \textbf{67.34} ± 6.57  & \textbf{55.18} ± 6.42  & \textbf{58.89} ± 13.65  &  \textbf{49.77} ± 14.72 \\
    \textsc{ToC-Qwen + Pause}      & \cmark  &  24.71 ± 3.91  &  18.68 ± 2.95  & 45.69 ± 14.72  &  35.95 ± 13.59  &   51.05 ± 17.89 & 42.20 ± 18.40  \\
    \bottomrule
  \end{tabular}
  }
\end{table*}

\begin{table}[t]
    \caption{Dataset statistics of the three datasets.}
    \label{tab:dataset_stats}
    \centering
    \resizebox{\linewidth}{!}{
        \begin{tabular}{lrrrr}
            \toprule
             \textbf{Dataset} & \textbf{Records} & \textbf{Duration} & \textbf{Speakers} & \textbf{Levels} \\
             \midrule
             \textsc{AMI} & 139 & 73h & - & 3\\
             \textsc{Videoaula} & 34 & 23h & 5 & 4\\
             \textsc{LectureDE} & 96 & 27h & 4 & 5\\
             \bottomrule
        \end{tabular}
    }
\end{table}

\subsection{Datasets}
\label{sec:datasets}

For our experiments, we employ three topic segmentation datasets and provide an overview in \autoref{tab:dataset_stats}.
The \textsc{AMI} corpus \cite{AMI-2005} comprises 171 English meetings with 139 sessions annotated for topic segmentation.
We adapt the topic interval annotations of \textsc{AMI} to our segmentation scheme (Section~\ref{sec:hier_metric}) by inserting a filler topic whenever a gap occurs. This preserves the practice of predicting only topic changes \cite{Retkowski-2024, Solbiati-2021, Koshorek-2018, Lukasik-2020}. To create a hierarchical structure, we treat any topic that begins within a larger segment as a subtopic. However, \textsc{AMI} is relatively shallow in hierarchical depth compared to the other two datasets.
The \textsc{Videoaula} dataset \cite{Soares-2018} contains 34 Portuguese video lectures on computer science topics. Each lecture is given by a distinct speaker and includes topic segment annotations.
\textsc{LectureDE}~\cite{freisinger2023_slate} is an internal collection of German video lectures covering engineering, social sciences, and didactics. 
We manually annotated these recordings with hierarchical topic labels to facilitate multi-level segmentation experiments.

For all datasets, we follow the same preprocessing procedure. 
We first generate transcripts with Whisper large-v2 \cite{Radford-2022}, an end-to-end multilingual ASR model trained on 680,000 hours of speech. 
Next, we apply the Montreal Forced Aligner \cite{McAuliffe-2017} to synchronize the transcripts with the respective audio tracks. 
Finally, we segment the aligned text into sentences using SpaCy\footnote{https://github.com/explosion/spaCy}.
This pipeline aims to approximate realistic conditions in which content is transcribed and segmented automatically.

\subsection{Hierarchical Evaluation}
\label{sec:hier_metric}
When evaluating methods producing hierarchical segmentation, standard single-level metrics (e.g., \(F_1\), \(B\), or \(P_k\)) lose the finer details or force one to select only a single level for evaluation. To address this, we propose a \emph{hierarchical metric wrapper} that yields a single numeric measure for multi-level segmentations, without discarding any coarse or fine levels.
We adopt a simple dynamic programming approach to align each reference level with a non-decreasing hypothesis level, reminiscent of monotonic sequence alignment methods (e.g., dynamic time warping), ensuring the reference’s coarse-to-fine progression matches the hypothesis in a globally optimal way.

Let \(D\) be a document with \(n\) sentences, and assume we have a \emph{reference} hierarchical segmentation \(\{S_1^l, \ldots, S_{m_l}^l\}\) at each level \(l \in \{1,\dots,L_{\text{ref}}\}\). Likewise, let \(\{H_1^k, \ldots, H_{p_k}^k\}\) be the \emph{hypothesis} segmentation at each level \(k \in \{1,\dots,L_{\text{hyp}}\}\). For each single level, we can compute a linear measure \(M(R^l,H^k)\), e.g., boundary similarity \(B\)~\cite{Fournier-2013}.

To define a single \emph{hierarchical} score, we require that if the reference progresses from a coarser level \(l\) to a finer level \(l+1\), the hypothesis does not move "back up" to a less-fine level. Formally, we choose a function \(\phi\) that satisfies:
\begin{align}
\phi : \{1,\dots,L_{\text{ref}}\} &\to \{1,\dots,L_{\text{hyp}}\}, 
\label{eq:phi-constraint}\\
\text{where} \quad
\phi(l+1) &\ge \phi(l) 
\quad \forall \, l.
\nonumber
\end{align}
Given this constraint, we seek the \(\phi^*\) that \emph{maximizes} the total boundary similarity:
\begin{align}
\phi^*
&=
\underset{\phi(\cdot)}{\mathrm{argmax}}
\sum_{l=1}^{L_{\text{ref}}}
B\bigl(R^l, H^{\phi(l)}\bigr),
\text{subject to}\
\phi(l+1)\ge \phi(l).
\label{eq:phi-max}
\end{align}
Finally, we define our hierarchical boundary similarity by averaging across all reference levels:
\begin{align}
B_{\text{hier}}(R,H)
&=
\frac{1}{L_{\text{ref}}}
\sum_{l=1}^{L_{\text{ref}}}
B\bigl(R^l, H^{\phi^*(l)}\bigr).
\label{eq:hier-bound}
\end{align}

Thus, each reference level \(l\) is matched to whichever hypothesis level maximizes its local boundary metric, while obeying the non-decreasing constraint of \eqref{eq:phi-constraint}.
Equations~\eqref{eq:phi-max} and \eqref{eq:hier-bound} then yield one scalar summarizing how well the entire \emph{nested} segmentation aligns, from coarse to fine levels.

\subsection{Baselines}
We compare our proposed method with several linear baselines to investigate their effectiveness in solving the topic segmentation task:

\textbf{\textsc{TT-BERT}}: As unsupervised baseline, we adopt the variant of TextTiling \cite{Hearst-1997} in conjunction with sentence embeddings \cite{Solbiati-2021}.
To compute the embeddings, we use a pre-trained text encoder\footnote{\textit{paraphrase-multilingual-mpnet-base-v2}} from the \textit{SentenceTransformers} library \cite{Reimers-2020}.

\textbf{\textsc{MiniSeg}}: As supervised baseline, we apply the sequence-tagging approach of \textsc{MiniSeg} \cite{Retkowski-2024}, which is based on MiniLM~\cite{Wang-2020} and RoFormer \cite{Su-2024}.
For each sentence, we determine whether a topic boundary is present or not.

\textbf{\textsc{CS-BERT}}: Another supervised baseline represents cross-segment BERT \cite{Lukasik-2020}, which includes a binary segment boundary classification. 
We concatenate the left and right context and use the \texttt{[CLS]} token for classification.

\textbf{\textsc{SegmentLLM}}: Finally, we adopt the prompt-based method introduced by \cite{Fan-2024}. 
This zero-shot approach instructs an LLM to generate a list of sentence indices, where each list represents a topic segment.

\subsection{Linear Segmentation Results}
Table~\ref{tab:results} summarizes the linear segmentation results, for which we use the most fine-grained segmentation level (cf. Section~\ref{sec:problem_definition}).
All supervised methods in Table~\ref{tab:results} are pre-trained on the first 10{,}000 articles of the \textsc{Wiki-727} corpus~\cite{Koshorek-2018}, similar to~\cite{Retkowski-2024, Ghosh-2022}. 
For \textsc{AMI}, we report the \textit{mean ± stddev} across 100 bootstrapped test sets, following~\cite{Lukasik-2020}. For \textsc{Videoaula} and \textsc{LectureDE}, we employ a \textit{leave-one-speaker-out} cross-validation strategy and report the \textit{mean ± stddev} over all validation sets. Metrics for each validation set are computed as the average across all samples in that set.

Among the fine-tuned models, \textsc{ToC-Nemo} with pause durations achieves the best performance across all datasets. This indicates that the model can leverage additional acoustic cues when supervised fine-tuning is applied. In contrast, pause durations lower the scores for the zero-shot variant, suggesting that these cues require specific training to yield improvements.

\textsc{SegmentLLM} with ChatGPT~\cite{Fan-2024} achieves the highest zero-shot scores. However, when both \textsc{SegmentLLM} and our ToC prompting rely on the Nemo model, the ToC approach consistently outperforms \textsc{SegmentLLM} across all datasets. This finding highlights the advantage of a table of contents format, even when only linear segmentation is evaluated. We did not run the \textsc{SegmentLLM ChatGPT} approach on \textsc{LectureDE}, because privacy restrictions prevent external cloud-based processing for this dataset.

\begin{figure}[t]
  \centering
  \includegraphics[width=\linewidth]{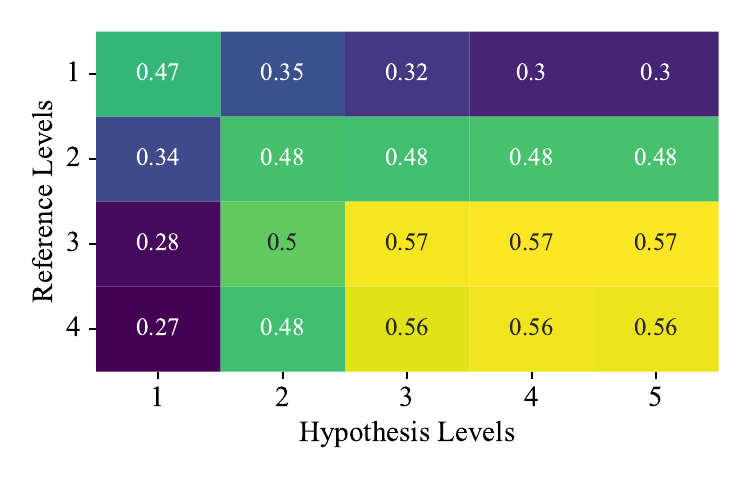}
  \caption{Level-specific $B$ scores for each reference (rows) to hypothesis (column) level combination. Average over all samples of \textsc{ToC-Nemo + Pause} on \textsc{Videoaula}}
  \label{fig:heat_levels}
\end{figure}

\renewcommand{\arraystretch}{1.1}
\begin{table}[t]
  \caption{Hierarchical $B$ scores (with difference to linear $B$ scores) for different 0-shot methods and datasets.}
  \label{tab:hier_results}
  \centering
  \resizebox{\linewidth}{!}{
  \begin{tabular}{l c c c}
    \toprule
    \textbf{Methods} & \textbf{\textsc{AMI}} & \textbf{\textsc{Videoaula}} & \textbf{\textsc{LectureDE}} \\
    \midrule
    \textsc{TT-BERT} &  6.70 (-0.01) & 8.64 (-0.09) &  7.92 (-0.39) \\
    \textsc{S.LLM ChatGPT}  &  8.49 (+0.52) &  20.2 (-2.09) & - \\
    \textsc{S.LLM Nemo}  & 5.77 (-0.02) &  4.17 (-0.51)  & 8.72 (-2.65)  \\
    \textsc{ToC-Nemo}  &   8.50 (+0.82)  & 14.01 (-0.05) &  21.56 (+4.52)  \\
    \textsc{ToC-Nemo + P.} & 8.10 (+1.05)  & 12.00 (-0.40)  &  21.73 (+5.94)  \\
    \bottomrule
  \end{tabular}
  }
\end{table}

\subsection{Hierarchical Segmentation Results}
\label{sec:hier_results}
The linear scores in Table~\ref{tab:results} do not reveal whether our methods truly produce nested structures, with coarse top-level and finer sub-level segments. To address this, we apply the extended boundary similarity metric (\(B_{hier}\)) introduced in Section~\ref{sec:hier_metric}.
This metric assesses the matching of the hypothesized and reference levels, thus reflecting the nested organization of topics.

Figure~\ref{fig:heat_levels} illustrates the \(B\)-scores across different reference and hypothesis levels for the \textsc{Videoaula} dataset, averaged over all lectures. The chosen method is \textsc{ToC-Nemo + Pause}. As the figure shows, the highest scores appear when comparing levels of similar granularity. This indicates that the model successfully captures both broad top-level and fine-grained lower-level topics, producing a meaningful table of contents structure.

Table~\ref{tab:hier_results} reports \(B_{hier}\) for the zero-shot methods from Table~\ref{tab:results}, along with the difference to the linear $B$ scores.
On the \textsc{LectureDE} dataset, which has deeper hierarchies (see Table~\ref{tab:dataset_stats}), the hierarchical metric for the linear-only segmenter \textsc{SegmentLLM Nemo} is lower than its standard linear \(B\)-scores.
Conversely, methods designed for multi-level output achieve higher \(B_{hier}\) values compared to their linear \(B\), underscoring the benefit of explicitly modeling nested topics.
For \textsc{AMI}, whose annotations are relatively shallow, the gap between \(B\) and \(B_{hier}\) is less pronounced across all methods.
These findings highlight the added structural value of hierarchical segmentation, indicating a need for evaluation metrics that go beyond simple linear boundaries.

\section{Conclusion}
We introduce a hierarchical topic segmentation approach that generates a multi-level table of contents from lengthy speech transcripts. Our experiments demonstrate that LLMs can surpass traditional baselines when fine-tuned. Including pause-duration cues provides an additional boost in segmentation performance. 
In addition, we propose a hierarchical evaluation method, offering deeper insights compared to linear metrics.

Nonetheless, zero-shot models still show substantial performance gaps, pointing to the need for more effective solutions that do not rely on supervised data from the target domain or language. We hope our findings inspire additional work on hierarchical segmentation, which can add valuable structure to automatically generated transcripts.

\bibliographystyle{IEEEtran}
\bibliography{mybib}

\end{document}